\pdfoutput=1

\documentclass[11pt]{article}

\usepackage[]{acl}

\usepackage{times}
\usepackage{latexsym}

\usepackage[T1]{fontenc}

\usepackage[utf8]{inputenc}

\usepackage{microtype}

\usepackage{graphicx}
\usepackage{picinpar}
\usepackage{subcaption}
\usepackage{colortbl}
\usepackage{booktabs}
\usepackage{multirow}
\usepackage{soul}
\usepackage{enumerate}
\usepackage{color}
\usepackage{hyperref}
\usepackage{amsmath}
\usepackage{amssymb}
\usepackage{algorithm}
\usepackage{algpseudocode}

\title{MentalBERT: \\Publicly Available Pretrained Language Models for Mental Healthcare}

\author{Shaoxiong Ji$^{\dag}$, Tianlin Zhang$^{\ddag}$, Luna Ansari$^{\dag}$, Jie Fu$^{\S}$, Prayag Tiwari$^{\dag}$, and Erik Cambria$^{\P}$\\ 
$^{\dag}$ Aalto University, Finland $^{\ddag}$ The University of Manchester, UK\\
$^{\S}$ Mila, Qu\'ebec AI Institute, Canada
$^{\P}$ Nanyang Technological University, Singapore\\
\texttt{\{shaoxiong.ji;~luna.ansari;~prayag.tiwari\}@aalto.fi} \\
\texttt{tianlin.zhang@postgrad.manchester.ac.uk} \\
\texttt{fujie@mila.quebec}~~~~~\texttt{cambria@ntu.edu.sg} \\
}

\begin{document}
\maketitle
\begin{abstract}
Mental health is a critical issue in modern society, and mental disorders could sometimes turn to suicidal ideation without adequate treatment. 
Early detection of mental disorders and suicidal ideation from social content provides a potential way for effective social intervention. 
Recent advances in pretrained contextualized language representations have promoted the development of several domain-specific pretrained models and facilitated several downstream applications. 
However, there are no existing pretrained language models for mental healthcare. 
This paper trains and release two pretrained masked language models, i.e., MentalBERT and MentalRoBERTa, to benefit machine learning for the mental healthcare research community. 
Besides, we evaluate our trained domain-specific models and several variants of pretrained language models on several mental disorder detection benchmarks and demonstrate that language representations pretrained in the target domain improve the performance of mental health detection tasks. 
\end{abstract}

\section{Introduction}
\label{sec:introduction}

Mental health is a global issue, especially severe in most developed countries and many emerging markets. 
According to the mental health action plan (2013 - 2020) from the World Health Organization, 1 in 4 people worldwide suffer from mental disorders to some extent. 
Moreover, 3 out of 4 people with severe mental disorders do not receive treatment, worsening the problem. 
During some periods like the pandemic, people struggle with mental health issues, and many may not get mental health practitioners' help. 
Previous studies reveal that suicide risk usually has a connection to mental disorders~\citep{windfuhr2011suicide}.
Partly due to severe mental disorders, 900,000 people commit suicide each year worldwide, making suicide the second most common cause of death among the young. 
Suicide attempters have been reported as suffering from mental disorders, with an investigation on a shift from mental health to suicidal ideation conducted by language and interactional measures~\citep{de2016discovering}.

Early identification is a practical approach to mental illness and suicidal ideation prevention.
Except for traditional proactive screening, social media is a good channel for mental health care.
Social media platforms such as Reddit and Twitter provide anonymous space for users to discuss stigmatic topics and self-report personal issues.
Social content from users who wrote about mental health issues and posted suicidal ideation has been widely used to study mental health issues~\citep[e.g.,][]{ji2018supervised,tadesse2019detection}.
Machine learning-based detection techniques can empower healthcare workers in early detection and assessment to take an action of proactive prevention.

Recent advances in deep learning facilitate the development of effective early detection methods~\citep{ji2021suicidal}. 
A new trend in natural language processing (NLP), contextualized pretrained language models, has attracted much attention for various text processing tasks. 
The seminal work on a pretrained language model called BERT~\citep{devlin2019bert} utilizes bidirectional transformer-based text encoders and trains the model on a large-scale corpus. 
With the success of BERT, several domain-specific pretrained language models for learning text representations have also been developed and released, such as biomedical BERT~\citep{lee2020biobert} and clinical BERT~\citep{alsentzer2019publicly,huang2019clinicalbert} for the biomedical and clinical domain, respectively. 

However, there are no pretrained language models customized for the domain of mental healthcare. 
Our paper trains and releases two representative bidirectional masked language models, i.e., BERT and RoBERTa~\citep{liu2019roberta}, with corpus collected from social forums for mental health discussion. 
The pretrained models in the mental health domain are dubbed MentalBERT and MentalRoBERTa.  
To our best knowledge, this work is the first to pre-train language models for mental healthcare.
Besides, we conduct a comprehensive evaluation on several mental health detection datasets with pretrained language models in different domains.
We release the pretrained MentalBERTs with Huggingface's model repository, available at \url{https://huggingface.co/mental}.

\section{Methods and Setup}
\label{sec:methods}
This section introduces the language model pretraining technique and the pretraining corpus we collected. 
We then present the downstream tasks that we aim to solve by fine-tuning the pretrained models, and describing the setup of language model fine-tuning. 
Note that we aim to provide publicly available pretrained text embeddings as language resources and evaluate the usability in downstream tasks rather than propose novel pretraining techniques.

\subsection{Language Model Pretraining}
We follow the standard pretraining protocols of BERT and RoBERTa with Huggingface's Transformers framework~\citep{wolf2020transformers}.
These two models work in a bidirectional manner, and we follow their mechanism and adopt the same loss of masked language modeling during pretraining. 
We use the base network architecture for both models.
The BERT model we use is base uncased, which is 12-layer, 768-hidden, and 12-heads and has 110M parameters.
For the pretraining of RoBERTa-based MentalBERT, we apply the dynamic masking mechanism that converges slightly slower than the static masking. 
Instead of the domain-specific pretraining~\citep{gu2020domain} that trains language models from scratch, we adopt the training scheme similar to the domain-adaptive pretraining~\citep{gururangan2020don} that continues the pretraining in specific downstream domains. 
Specifically, we start the training of language models from the checkpoint of original BERT and RoBERTa. 
In this way, we can utilize the learned knowledge from the general domain and save computing resources, and continued pretraining makes the model adaptive to the target domain of mental health. 

We use four Nvidia Tesla v100 GPUs to train the two language models. 
The computing resources are also one of the main assets of this paper. 
We set the batch size to 16 per GPU, evaluate every 1,000 steps, and train for 624,000 iterations. 
Training with four GPUs takes around eight days, i.e., around 32 days with only one GPU.

\subsection{Pretraining Corpus}
We collect our pretraining corpus from Reddit, an anonymous network of communities for discussion among people of similar interests. 
Focusing on the mental health domain, we select several relevant subreddits (i.e., Reddit communities that have a specific topic of interest) and crawl the users' posts. 
We do not collect user profiles when collecting the pretraining corpus, even though those profiles are publicly available. 
The selected mental health-related subreddits include ``r/depression'', ``r/SuicideWatch'', ``r/Anxiety'', ``r/offmychest'', ``r/bipolar'', ``r/mentalillness/'', and ``r/mentalhealth''.
Eventually, we make the training corpus with a total of 13,671,785 sentences.

\subsection{Downstream Task Fine-tuning}
We apply the pretrained MentalBERT and MentalRoBERTa in binary mental disorder detection and multi-class mental disorder classification of various mental disorders such as stress, anxiety, and depression.
We fine-tune the language models in downstream tasks.
Specifically, we use the embedding of the special token \texttt{[CLS]} of the last hidden layer as the final feature of the input text.
We adopt the multilayer perceptron (MLP) with the hyperbolic tangent activation function for the classification model.
We set the learning rate of the transformer text encoder to be 1e-05 and the learning rate of classification layers to be 3e-05. 
The optimizer is Adam~\citep{kingma2014adam}.

\section{Results}
\label{sec:experiments}

\begin{table*}[htbp]
\small
\centering
\setlength{\tabcolsep}{6pt}
\caption{A summary of datasets. Note we hold out a portion of original training set as the validation set if the original dataset does not contain a validation set.}
\begin{center}
\begin{tabular}{lllrrr}
\toprule
Category	&	Platform	&	Dataset		&	train	&	validation	&	test	\\
\midrule
Assorted	&	Reddit	&	SWMH	\citep{ji2021suicidal}	&	34,823	&	8,706	&	10,883	\\
Depression	&	Reddit	&	eRisk18 T1	\citep{losada2016test}	&	1,533	&	658	&	619	\\
Depression	&	Reddit	&	Depression$\_$Reddit	\citep{pirina2018identifying}	&	1,004	&	431	&	406	\\
Depression	&	Reddit	&	CLPsych15	\citep{coppersmith2015clpsych}	&	457	&	197	&	300	\\
Stress	&	Reddit	&	Dreaddit	\citep{turcan2019dreaddit}	&	2,270	&	568	&	715	\\
Suicide	&	Reddit	&	UMD	\citep{shing2018expert}	&	993	&	249	&	490	\\
Suicide	&	Twitter	&	T-SID	\citep{ji2021suicidal}	&	3,072	&	768	&	960	\\
Stress	&	SMS-like	&	SAD	\citep{mauriello2021sad}	&	5,548	&	617	&	685	\\
\bottomrule
\end{tabular}
\end{center}
\label{tab:data}
\end{table*}%

\subsection{Datasets}
We evaluate and compare mental disorder detection methods on different datasets with various mental disorders (e.g., depression, anxiety, and suicidal ideation) collected from popular social platforms (e.g., Reddit and Twitter). 
Table~\ref{tab:data} summarizes the datasets used in this paper. 
We carefully choose those benchmarks to cover a relatively wide range of mental health categories and social platforms.
Some datasets do not provide a validation set.
Thus, we partition a small set from the original training set to make the validation set. 

\paragraph{Depression}
Depression is one of the most common mental disorders discussed on many social platforms. 
We take it as a representative to evaluate the performance of different pretrained models.
The first dataset for depression comes from the CLPsych 2015 Shared Task~\citep{coppersmith2015clpsych}\footnote{\url{http://www.cs.jhu.edu/~mdredze/datasets/clpsych_shared_task_2015/}}. 
The first task of CLPsych 2015 contains user-generated posts from users with depression on Twitter. 
The train partition consists of 327 depression users, and the test data contains 150 depression users.
Note that there is an unknown data missing issue in the dataset of the CLPsych 2015 shared task.
The second dataset used is from eRisk shared task 1~\citep{losada2016test}, which is a public competition for early risk detection in health-related areas. 
The eRisk dataset contains posts from 2,810 users, where 1,370 users express depression in their posts and 1,440 act as the control group without depression. 

\citet{pirina2018identifying} collected additional social data form Reddit and combined them with previously collected data to identify depression\footnote{\url{https://github.com/Inusette/Identifying-depression}}.
We term this dataset as \textit{Depression\_Reddit} in this paper.

\paragraph{Suicidal Ideation}
We use data collected from Reddit and Twitter to test the performance.
Firstly, we use the UMD Reddit Suicidality Dataset~\citep{shing2018expert} that has a total of 865 users in the subreddit of ``SuicideWatch'' in Reddit\footnote{\url{http://users.umiacs.umd.edu/~resnik/umd_reddit_suicidality_dataset.html}}. 
The raw data annotation labels the user posts with four levels of risks. 
We include the control users and transform the label space into three classes according to the level of risks.
In addition to data from Reddit, we also evaluate the performance of data collected from Twitter. 
We use the Twitter dataset with tweets expressing suicidal ideation and normal posts as the control group, which is collected by~\citet{ji2018supervised, ji2021suicidal}.
We term this dataset as T-SID.

\paragraph{Other Mental Disorders}
We also evaluate the performance of classifying other mental disorders such as stress, anxiety, and bipolar.
Dreaddit~\citep{turcan2019dreaddit} is a dataset for stress analysis with posts collected from five different forums of Reddit\footnote{\url{http://www.cs.columbia.edu/~eturcan/data/dreaddit.zip}}. 
Specifically, it considers three major stressful topics, i.e., interpersonal conflict, mental illness, and financial need, and collects posts from ten related subreddits, including some mental health domains such as anxiety and PTSD.
This dataset consists of a total of 3,553 posts split into train and test sets.
Many factors may cause stress. 
We then use another dataset for recognizing everyday stressors called SAD, which contains 6,850 SMS-like sentences~\citep{mauriello2021sad}.
The SAD dataset derives nine stress factors from stress management articles, chatbot-based conversation systems, crowdsourcing, and web crawling. 
The specific stressor categories include work, health, fatigue, or physical pain, financial problem, emotional turmoil, school, everyday decision making, family issues, social relationships, and other unspecified stressors.
Lastly, we use a dataset called SWMH~\citep{ji2021suicidal} that contains Reddit posts with various mental disorders, including stress, anxiety, bipolar, depression, and suicidal ideation. 
Note that this dataset uses weak labels during the annotation process.

\subsection{Baselines}
\label{sec:baseline}
We compare our pretrained language models for mental health with various existing pretrained models in different domains.
They are BERT and RoBERTa pretrained with general corpus, BioBERT pretrained in the biomedical domain, and ClinicalBERT pretrained with clinical notes.
Note that the aim of this paper is not to achieve the state-of-the-art performance but to demonstrate the usability and evaluate the performance of our pretrained models, though we have achieved competitive performance in some datasets when compared with the state of the art.

\begin{table*}[htbp!]
\small
\centering
\setlength{\tabcolsep}{6pt}
\caption{Results of depression detection. The bold text represents for the best performance.}
\begin{center}
\begin{tabular}{l|rr|rr|rr}
\toprule
\multirow{2}{4em}{Model}	&	\multicolumn{2}{c|}{eRisk T1}			&	\multicolumn{2}{c|}{CLPsych}			&	\multicolumn{2}{c}{Depression\_Reddit}			\\
	&	Rec.	&	F1	&	Rec.	&	F1	&	Rec.	&	F1	\\
\midrule
BERT	&	88.53	&	88.54	&	64.67	&	62.75	&	91.13	&	90.90	\\										
RoBERTa	&	92.25	&	92.25	&	67.67	&	66.07	&	\textbf{95.07}	&	\textbf{95.11}	\\
BioBERT	&	79.16	&	78.86	&	65.67	&	65.50	&	91.13	&	90.98	\\
ClinicalBERT	&	76.25	&	75.41	&	65.67	&	65.30	&	89.41	&	89.03	\\
\hline
MentalBERT	&	86.27	&	86.20	&	64.67	&	62.63	&	94.58	&	94.62	\\
MentalRoBERTa	&	\textbf{93.38}	&	\textbf{93.38}	&	\textbf{70.33}	&	\textbf{69.71}	&	94.33	&	94.23	\\
\bottomrule
\end{tabular}
\end{center}
\label{tab:results_depression}
\end{table*}%

\begin{table*}[htbp!]
\small
\centering
\setlength{\tabcolsep}{4pt}
\caption{Results of classifying other mental disorders including stress, anorexia, suicidal ideation. The bold text represents for the best performance.}
\begin{center}
\begin{tabular}{l|rr|rr|rr|rr|rr}
\toprule
\multirow{2}{4em}{Model}	&	\multicolumn{2}{c|}{UMD}			&	\multicolumn{2}{c|}{T-SID}			&	\multicolumn{2}{c|}{SWMH}			&	\multicolumn{2}{c|}{SAD}			&	\multicolumn{2}{c}{Dreaddit}			\\
	&	Rec.	&	F1	&	Rec.	&	F1	&	Rec.	&	F1	&	Rec.	&	F1	&	Rec.	&	F1	\\
\midrule
BERT	&	61.63	&	58.01	&	88.44	&	88.51	&	69.78	&	70.46	&	62.77	&	62.72	&	78.46	&	78.26	\\
RoBERTa	&	59.39	&	\textbf{60.26}	&	88.75	&	88.76	&	\textbf{70.89}	&	72.03	&	66.86	&	67.53	&	80.56	&	80.56	\\
BioBERT	&	57.76	&	58.76	&	86.25	&	86.12	&	67.10	&	68.60	&	66.72	&	66.71	&	75.52	&	74.76	\\
ClinicalBERT	&	58.78	&	58.74	&	85.31	&	85.39	&	67.05	&	68.16	&	62.34	&	61.25	&	76.36	&	76.25	\\
\hline
MentalBERT	&	\textbf{64.08}	&	58.26	&	88.65	&	88.61	&	69.87	&	71.11	&	67.45	&	67.34	&	80.28	&	80.04	\\
MentalRoBERTa	&	57.96	&	58.58	&	\textbf{88.96}	&	\textbf{89.01}	&	70.65	&	\textbf{72.16}	&	\textbf{68.61}	&	\textbf{68.44}	&	\textbf{81.82}	&	\textbf{81.76}	\\
\bottomrule
\end{tabular}
\end{center}
\label{tab:results_others}
\end{table*}%

\subsection{Results and Discussion}
\label{sec:results}
We evaluate the model performance by comparing the recall and F1 scores. 
Mental disorder detection is usually a task with unbalanced classes, leading to using the F1 score metric.
It is also essential to reduce the false negatives, i.e., to ensure as few cases as possible that the detection model misses people with mental disorders.
Thus, we also report recall scores.

\paragraph{Results of Depression Detection}
We first compare the performance of depression detection.
Table~\ref{tab:results_depression} reports the results on three depression dataset collected from Reddit.
MentalRoBERTa archives the best performance on eRisk and CLPsych datasets, and MentalBERT is the second best model on the Depression$\_$Reddit dataset.

\paragraph{Results of Classifying Other Mental Disorders}
We then compare the performance of classifying other mental disorders and suicidal ideation. 
Table~\ref{tab:results_others} shows the performance on various datasets with different mental disorder classification tasks.
In T-SID, SAD, and Dreaddit, MentalRoBERTa is the best model with the highest recall and F1 scores. 
The MentalBERT has the highest F1 score in the UMD dataset, while its F1 score is not competitive to other models. 
While for the SWMH dataset with several mental disorders, the MentalRoBERT obtained the best F1 score.

\paragraph{Discussion}
When comparing the domain-specific pretrained models for mental health with models pretrained with general corpora, MentalBERT and MentalRoBERTa gain better performance in most cases. 
Domain-specific pretraining in the biomedical or clinical domain turns out to be less helpful than pretraining on the target domain of mental health. 
Those results show that continued pretraining on the mental health domain improves prediction performance in downstream tasks of mental health classification.

\section{Related Work}
\label{sec:related}

\paragraph{Contextualized Text Embeddings}
Contextualized embeddings have been intensively studied in NLP. 
Self-supervised large-scale pretraining facilitates the learning of semantic and contextual information and benefits various downstream applications such as text classification~\citep{sun2019fine}, sentiment analysis~\citep{tang2020fine,song2020utilizing} and relation extraction~\citep{alt2019fine}.
There are also many domain-specific variants of pretrained contextualized text embeddings.
Embeddings in specific domains aim to encode domain-specific information to boost the performance of a specific domain. 
For example, BioBERT~\citep{lee2020biobert} pretrained the BERT model in the biomedical domain using research articles from PubMed, which was applied to many biomedical tasks such as biomedical relation extraction and named entity recognition.  
ClinicalBERT~\citep{alsentzer2019publicly} used clinical notes as the pretraining corpus to continue the pretraining of the BERT model. 
Those domain-specific variants also foster variable downstream applications by fine-tuning pretrained embeddings such as~\citet{lin2019bert} and~\citet{ji2020bert}.

\paragraph{NLP for Mental Healthcare}

Mental healthcare research in social media is increasingly applying NLP techniques to capture users' behavioral tendencies. 
Various methods are implemented for labeling, i.e., identifying emotions, mood, and profiles that might indicate mental health problems~\cite{calvo2017natural}. 
One of the most representative tasks is mental health detection that categorizes given social posts into different classes of mental disorders such as depression~\citep{tadesse2019detection}.
Mental state understanding requires effective feature representation learning and complex emotive processes. 
\citet{resnik2013using} applied topic modeling, an unsupervised approach that reduces the input of textual data feature space to a fixed number of topics to feature engineering in depression detection. 
Feature engineering-based machine learning method designs manual features and builds classifiers for mental health detection~\citep{shatte2019machine,abd2020application}. 
Various features such as sensor signals from personal devices~\citep{mohr2017personal} and EEG signals~\citep{gore2019surveying} have been applied. 
For detection from textual data in particular, text features include word counts, TF-IDF~\citep{campillo2021nlp}, topic features~\citep{shickel2020automatic} and sentiment traits~\citep{yoo2019semantic}.
Severe mental disorders but without intervention may lead to suicidal ideation~\citep{windfuhr2011suicide}. 
Many machine learning-based methods have been applied for suicidal ideation detection~\citep{ji2021sid}. 

Recent work applies deep representation learning methods, which enable automatic feature learning to solve the early mental disorder identification task.
Those methods typically build text embeddings and feed the embeddings into neural architectures such as convolutional neural networks~\citep{rao2020mgl}, recurrent networks~\citep{bouarara2021recurrent}, self attention-based Transformers, hybrid architectures like CNN-LSTM~\citep{kang2021classification} and more other deep learning architectures~\citep{su2020deep}.
Recent works, e.g.,~\citet{jiang2020detection},~\citet{martinez2021bert} and~\citet{bucur2021early}, use pretrained language models and fine-tune the model for mental health tasks. 
However, there are no existing pretrained language models trained with mental health-related text to benefit the domain application directly.

\section{Conclusion and Future Work}
\label{sec:conclusion}
This paper trains and releases two masked language models, i.e., MentalBERT and MentalRoBERTa, on the domain data of mental health collected from the Reddit social platform. 
This paper is the first work that trains domain-specific language models for mental healthcare.
Our pretrained models are publicly available and can be reused by the research community.
Besides, we conduct a comprehensive evaluation on the performance for downstream mental health detection tasks, including depression, stress, and suicidal ideation detection. 
Our empirical results show that continued pretraining with mental health-related corpus can improve classification performance.

Our paper is a positive attempt to benefit the research community by releasing the pretrained models for other practical studies and with the hope to facilitate some possible real-world applications to relieve people's mental health issues. 
However, we only focus on the English language in this study since English corpora are relatively easy to obtain.
In the future work, we plan to collect multilingual mental health-related posts, especially those less studied by the research community, and train a multilingual language model to benefit more people speaking languages other than English.

\section*{Social Impact}
The paper trains and releases masked language models for mental health to facilitate the automatic detection of mental disorders in online social content for non-clinical use. 
The models may help social workers find potential individuals in need of early prevention. 
However, the model predictions are not psychiatric diagnoses. 
We recommend anyone who suffers from mental health issues to call the local mental health helpline and seek professional help if possible.

Data privacy is an important issue, and we try to minimize the privacy impact when using social posts for model training.
During the data collection process, we only use anonymous posts that are manifestly available to the public. 
We do not collect user profiles even though they are also manifestly public online. 
We have not attempted to identify the anonymous users or interact with any anonymous users. 
The collected data are stored securely with password protection even though they are collected from the open web.
There might also be some bias, fairness, uncertainty, and interpretability issues during the data collection and model training. 
Evaluation of those issues is essential in future research. 

\section*{Acknowledgments}
The authors would like to thank Philip Resnik for providing the UMD Reddit Suicidality Dataset, Mark Dredze for providing the dataset in the CLPsych 2015 shared task, and other researchers who make their datasets publicly available.
We acknowledge the computational resources provided by the Aalto Science-IT project.
The authors wish to acknowledge CSC - IT Center for Science, Finland, for computational resources. 

\bibliography{MentalBERT.bbl}
\bibliographystyle{acl_natbib}

\end{document}